\documentclass[nohyperref]{article}
\usepackage{microtype}
\usepackage{graphicx}
\usepackage{subfigure}
\usepackage{booktabs} 
\usepackage{multicol}
\usepackage{multirow}
\usepackage{hyperref}
\usepackage{natbib}

\usepackage[accepted]{icml2022}
\usepackage{amsmath}
\usepackage{amssymb}
\usepackage{mathtools}
\usepackage{amsthm}

\theoremstyle{plain}

\theoremstyle{definition}

\theoremstyle{remark}

\usepackage[textsize=tiny]{todonotes}

\icmltitlerunning{Formatting of ICML 2022}

\begin{document}

\twocolumn[
\icmltitle{Light Aircraft Game : Basic Implementation and training results analysis}



\icmlsetsymbol{equal}{*}

\begin{icmlauthorlist}
\icmlauthor{Hanzhong Cao}{yyy}
\end{icmlauthorlist}

\icmlaffiliation{yyy}{Department of Computer Science, Peking University, Beijing, China}

\icmlcorrespondingauthor{Hanzhong Cao}{2200012929@stu.pku.edu.cn}

\icmlkeywords{Machine Learning, ICML}

\vskip 0.3in
]



\printAffiliationsAndNotice{}  

\begin{abstract}
This paper investigates multi-agent reinforcement learning (MARL) in a partially observable, cooperative-competitive combat environment known as LAG. We detail the environment's settings, including agent actions, hierarchical controls, and reward design across different combat modes such as \textit{No Weapon} and \textit{ShootMissile}. Two representative algorithms are evaluated: HAPPO, an on-policy hierarchical variant of PPO, and HASAC, an off-policy method based on soft actor-critic. We analyze their training stability, reward evolution, and inter-agent coordination abilities. Experimental results demonstrate that HASAC excels in simpler coordination tasks without weapons, while HAPPO shows superior adaptability in more dynamic and expressive scenarios like missile combat. These findings offer insight into the trade-offs between on-policy and off-policy methods in multi-agent settings. \href{https://github.com/xiaxiaoguang/LightAircraftGame}{Our Github Repo}

\end{abstract}

\section{Introduction}


In recent years, \textbf{Multi-Agent Reinforcement Learning (MARL)} has emerged as a critical area of research within the broader field of machine learning. While traditional reinforcement learning focuses on training a single agent to interact with an environment and maximize cumulative rewards, MARL extends this paradigm to multiple agents that learn and act simultaneously, often in cooperative or competitive settings. This added complexity introduces challenges such as non‑stationarity, partial observability, and the need for decentralized coordination. Unlike single‑agent methods, MARL requires policies that can handle dynamic interactions among agents, which may lead to emergent behaviors that are difficult to predict or model. As a result, MARL research has increasingly focused on scalable architectures, stability under multi‑agent dynamics, and policy generalization.

With the advancement of MARL algorithms, a wide variety of benchmark environments have been created to evaluate different aspects of multi‑agent learning. In continuous robotic control, platforms such as DexHand\cite{chen2022humanlevelbimanualdexterousmanipulation} and MA‑MuJoCo\cite{peng2021facmacfactoredmultiagentcentralised} enable high‑fidelity simulation of dexterous manipulation and physics‑based locomotion. In the realm of team sports and strategy, Google Research Football provides a rich, physics‑driven soccer environment requiring tactical cooperation. Light Aircraft Game (LAG) is a lightweight, scalable, Gym‑wrapped aircraft combat environment designed for rapid experimentation in aerial dogfights and team skirmishes. PettingZoo offers a simple, Pythonic interface for representing general MARL problems, complete with both the Agent Environment Cycle (AEC) API for sequential turn‑based tasks and the Parallel API for simultaneous action settings, plus a suite of reference environments and utilities for custom development. Finally, SMAC\cite{samvelyan2019starcraftmultiagentchallenge} (StarCraft Multi‑Agent Challenge) and its successor SMACv2\cite{ellis2023smacv2improvedbenchmarkcooperative} present agents with partial observations in real‑time combat inspired by StarCraft II. Together, these environments span the spectrum from continuous control to discrete strategic games, providing diverse testbeds for MARL research.

This report focuses specifically on the \textbf{Light Aircraft Game (LAG)} environment. LAG is a competitive, aircraft‑themed Gym environment that emphasizes \emph{lightweight deployment}, \emph{scalability to many concurrent agents}, and \emph{easy customization of reward structures}. Agents pilot simple fighter aircraft in configurable scenarios—ranging from free‑for‑all skirmishes to coordinated team engagements—and must learn both low‑level control (e.g., throttle, pitch, yaw) and high‑level tactics (e.g., formation flying, target prioritization). In our study, we select five representative LAG tasks, including \emph{2v2 Noweapon} (unarmed dogfight) and \emph{Shoot Missile War} (armed engagement), to evaluate how well modern MARL algorithms generalize across different aerial combat scenarios.

Our investigation centers on two MARL algorithms: \textbf{HAPPO} (Heterogeneous‑Agent Proximal Policy Optimization)\cite{kuba2022trustregionpolicyoptimisation} and \textbf{HASAC} (Heterogeneous‑Agent Soft Actor‑Critic)\cite{liu2025maximumentropyheterogeneousagentreinforcement}. HAPPO extends PPO to multi‑agent settings by enforcing trust‑region updates in a decentralized manner, while HASAC incorporates entropy‑regularized objectives to balance exploration and stability. By training both methods across the five LAG tasks, we analyze their learning curves, convergence properties, and emergent behaviors. The remainder of this report details our experimental setup, presents quantitative training results, and offers a comparative analysis of the two algorithms' performance in the Light Aircraft Game.

\section{Preliminary}
\label{Preliminary}

\subsection{Simulation Framework}

The LAG environment is constructed around two principal simulation components: the \textit{Aircraft Simulator} and the \textit{Missile Simulator}.

\paragraph{Aircraft Simulator.} This module models the aircraft's flight dynamics. It determines the current position using variables such as $\mathit{delta\_altitude}$, $\mathit{altitude}$, $\mathit{longitude}$, $\mathit{latitude}$, and $\mathit{delta\_heading}$. It also records internal state parameters including roll, pitch, three-dimensional velocity $(v_x, v_y, v_z)$, and acceleration $(a_x, a_y, a_z)$. These values enable accurate prediction of the aircraft's future trajectory.

\paragraph{Missile Simulator.} Building on the physical model of the aircraft, this module incorporates additional aspects relevant to missile behavior, such as aerodynamic drag, explosive radius, and missile lifespan. A proportional navigation guidance system is employed for realistic missile trajectory control.

\paragraph{Control Interface.} The agent interacts with the aircraft through four continuous control inputs: aileron (roll), elevator (pitch), rudder (yaw), and throttle (thrust). To address the complexity of joint flight and combat behavior, we adopt a hierarchical control paradigm. The high-level controller sets targets for direction, altitude, and velocity, while the low-level controller—trained via the SingleControl task—executes fine-grained actuation commands.

\subsection{Task Suite}

The LAG environment encompasses three progressively complex task categories: \textit{SingleControl}, \textit{SingleCombat}, and \textit{DualCombat}.

\paragraph{SingleControl.} This task is designed to train the low-level controller to stabilize and maneuver the aircraft effectively. It serves as a foundational module for downstream combat tasks.

\paragraph{SingleCombat.} This category includes 1-vs-1 aerial engagements between two aircraft agents. Two distinct sub-tasks are provided:
\begin{itemize}
    \item \textbf{NoWeapon Task.} Inspired by reconnaissance operations, the agent must maintain a positional advantage by maneuvering behind the opponent while preserving a safe and controlled distance.
    \item \textbf{Missile Task.} The agent is required not only to maneuver but also to engage in missile combat. This task is further divided into:
    \begin{itemize}
        \item \textit{Dodge Missile.} Missile launches follow predefined rules. The agent must learn to evade incoming missiles.
        \item \textit{Shoot Missile.} Missile launching becomes a learning objective. Since training from scratch is challenging due to sparse rewards, we incorporate prior knowledge using the conjugate property of the Beta distribution for binomial processes. This probabilistic prior aids in policy learning for missile firing decisions.
    \end{itemize}
\end{itemize}
Both sub-tasks support self-play and agent-vs-baseline training settings.

\paragraph{DualCombat.} In this cooperative-competitive setting, each team controls two aircraft. The goal remains consistent with the SingleCombat tasks, but with additional requirements for intra-team coordination. The high-level strategy module plays a crucial role in orchestrating team behavior during engagement.




\subsection{Reward Design}

In our multi-agent dual-aircraft combat experiments, the reward function plays a critical role in shaping agent behavior. The LAG environment employs a composite reward mechanism comprising three categories: \textit{AltitudeReward}, \textit{PostureReward}, and \textit{EventDrivenReward}. Each reward type captures distinct aspects of tactical air combat performance and safety.

\paragraph{AltitudeReward.} This component penalizes unsafe flight behavior, particularly when the aircraft violates minimum altitude constraints. It is defined as:
\begin{itemize}
    \item \textbf{Velocity Penalty:} A negative reward is assigned when the aircraft's velocity is insufficient while flying below a safe altitude. The typical reward range is $[-1, 0]$.
    \item \textbf{Altitude Penalty:} Additional penalty is applied when the aircraft descends below a danger altitude threshold. This discourages risky low-altitude flight and enforces adherence to operational constraints. The reward range is likewise $[-1, 0]$.
\end{itemize}

\paragraph{PostureReward.} This term encourages advantageous spatial and directional alignment between the agent and its opponent. It is modeled as the product of two factors:
\begin{itemize}
    \item \textbf{Orientation:} Positive reward is given when the agent aligns its heading toward the enemy fighter. Conversely, being targeted by the opponent incurs a penalty.
    \item \textbf{Range:} Agents are rewarded for maintaining proximity to the enemy within an effective engagement zone, while excessive distance results in negative feedback.
\end{itemize}

\paragraph{EventDrivenReward.} This sparse, high-magnitude reward is triggered by critical events during the combat engagement:
\begin{itemize}
    \item \textbf{Shot Down by Missile:} $-200$ reward is assigned to penalize being destroyed by an enemy missile.
    \item \textbf{Crash:} Accidental crashes due to poor control or environmental factors also incur a $-200$ penalty.
    \item \textbf{Enemy Kill:} Successfully shooting down an opponent yields a substantial reward of $+200$.
\end{itemize}

Together, these reward components form a balanced and hierarchical structure that guides learning from low-level flight safety to high-level combat effectiveness. The design enables agents to gradually acquire safe, stable, and strategically advantageous behavior in both cooperative and adversarial scenarios.

\section{Methodology}
\label{Methodology}

\subsection{Heterogeneous-Agent Proximal Policy Optimization (HAPPO)}

In the context of multi-agent reinforcement learning (MARL), achieving stable and monotonic policy improvement presents a major challenge due to the inherently non-stationary and interdependent nature of agent interactions. Even in cooperative settings, agents may induce conflicting policy updates, undermining joint performance. To address this, the \textit{Heterogeneous-Agent Proximal Policy Optimization} (HAPPO) algorithm extends the trust region learning framework to MARL, enabling agents to optimize their individual policies while maintaining a principled guarantee of joint policy improvement.

HAPPO is founded on two theoretical pillars: the \textit{multi-agent advantage decomposition lemma} and a \textit{sequential policy update scheme}. These allow the policy of each agent to be optimized one at a time, while accounting for the potential influence of prior agent updates. Unlike traditional MARL algorithms that assume parameter sharing or require decomposition of the joint value function, HAPPO makes no such restrictive assumptions. It enables decentralized learning by allowing each agent to learn an individual policy, thereby improving scalability and generality across heterogeneous agent settings.

The core objective of HAPPO is a clipped surrogate loss function extended to the multi-agent case. The update for agent $i_m$ is computed as:

\begin{align}
&\mathbb{E}_{s \sim p_{\pi_{\theta_k}},\, a \sim \pi_{\theta_k}} \Bigg[
    \min \Bigg(
        \frac{\pi^{i_m}_{\theta^{i_m}}(a^{i_m} \mid s)}{\pi^{i_m}_{\theta^{i_m}_k}(a^{i_m} \mid s)} 
        M^{i_{1:m}}(s, a), \nonumber \\
&\hspace{4em} 
        \mathrm{clip}\left(
            \frac{\pi^{i_m}_{\theta^{i_m}}(a^{i_m} \mid s)}{\pi^{i_m}_{\theta^{i_m}_k}(a^{i_m} \mid s)}, 
            1 \pm \epsilon
        \right) 
        M^{i_{1:m}}(s, a)
    \Bigg)
\Bigg]
\end{align}

where $M^{i_{1:m}}(s, a)$ serves as a multi-agent modification factor for the advantage function, defined by:

\begin{equation}
M^{i_{1:m}}(s, a) = \frac{\hat{\pi}^{i_{1:m-1}}(a^{i_{1:m-1}} | s)}{\pi^{i_{1:m-1}}(a^{i_{1:m-1}} | s)} A(s, a),
\end{equation}

with $\hat{\pi}^{i_{1:m-1}}$ denoting the updated policies of previous agents in the sequence, and $A(s, a)$ representing the advantage of the joint action $a$ in state $s$.

\subsection{Heterogeneous-Agent Soft Actor-Critic (HASAC)}

Heterogeneous-Agent Soft Actor-Critic (HASAC) is a multi-agent reinforcement learning (MARL) algorithm developed to address critical limitations in existing methods, such as poor sample efficiency, unstable training dynamics, and convergence to suboptimal Nash equilibria in cooperative tasks. HASAC is derived by embedding cooperative MARL settings into probabilistic graphical models and adopting the Maximum Entropy (MaxEnt) reinforcement learning framework, which encourages agents to act stochastically and explore effectively.

By maximizing both the expected cumulative reward and policy entropy, HASAC fosters more diverse and stable policies, especially valuable in environments requiring sustained exploration and resilience to policy fluctuation. The algorithm extends the Soft Actor-Critic (SAC) approach to multi-agent scenarios with heterogeneous agents, each maintaining its own actor and critic networks. Importantly, HASAC allows agents to learn independently while still optimizing a globally cooperative objective.

From a theoretical standpoint, HASAC enjoys two key guarantees: (1) monotonic improvement in policy updates, and (2) convergence to the \emph{quantal response equilibrium} (QRE), a relaxed form of Nash equilibrium that better accounts for stochastic decision-making. These properties are established through a unified framework called \textbf{Maximum Entropy Heterogeneous-Agent Mirror Learning (MEHAML)}, which generalizes the algorithmic design of HASAC and ensures that any derived method from this template inherits the same theoretical guarantees.

\section{Experiments}
\label{exp}

\section{Analysis}
\label{analysis}

\subsection{Full Results}

\subsection{Comparative Analysis of HAPPO and HASAC}

We evaluate HAPPO and HASAC under two gameplay conditions—\textit{No Weapon} and \textit{ShootMissile}—each comprising the HierarchySelfplay, SelfPlay, and vsBaseline evaluation protocols (Table~\ref{tab:happo_hasac_results}).

\subsubsection{No Weapon Setting}
In the \textit{No Weapon} scenario, HASAC consistently outperforms HAPPO across all protocols and timesteps:
\begin{itemize}
  \item \textbf{HierarchySelfplay:} HAPPO's returns remain negative throughout training (from –54.13 to –66.92), whereas HASAC maintains positive rewards (around 30).
  \item \textbf{SelfPlay:} HAPPO exhibits performance collapse (–13.81 to –66.92), while HASAC progresses from –24.88 to +7.82.
  \item \textbf{vsBaseline:} HAPPO is unable to exceed –95, in contrast to HASAC's stable near 30 reward.
\end{itemize}
These results indicate that the on‐policy nature of HAPPO struggles to stabilize coordination among multiple agents in purely positional tasks, whereas the off‐policy SAC foundation of HASAC delivers greater sample efficiency and robustness under limited action complexity.

\subsubsection{ShootMissile Setting}
Under the \textit{ShootMissile} condition, HAPPO demonstrates a marked advantage:
\begin{itemize}
  \item \textbf{HierarchySelfplay \& vsBaseline:} HAPPO's reward surges from 385.27 to over 1\,090.17, reflecting its capacity to learn expressive, high‐variance policies for missile engagement.
  \item \textbf{HASAC Performance:} Although HASAC improves (–6.79 to 735.59 in HierarchySelfplay; 4.77 to 468.34 in vsBaseline), it remains below HAPPO's peak.
\end{itemize}
This reversal suggests that HAPPO's clipped surrogate objective better supports exploration in high‐dimensional action spaces, enabling effective missile‐firing strategies, whereas HASAC's entropy‐regularized updates provide less aggressive policy refinement in these tasks.

\subsection{Algorithmic Trends Across Tasks}

\subsubsection{HAPPO: Variance and Task Dependence}
HAPPO exhibits:
\begin{itemize}
  \item \emph{High variance} in reward trajectories, particularly under decentralized training (SelfPlay) in the No Weapon setting.This aligns with known issues of gradient instability in on-policy methods such as PPO when agents are non-stationary and influence each other’s learning dynamics.\cite{dewitt2020independentlearningneedstarcraft,jiang2017deepreinforcementlearningframework}
  \item \emph{Strong expressiveness} when flexible, temporally extended behaviors (missile launch and evasion) are required.
\end{itemize}
These patterns imply that HAPPO's on‐policy updates are sensitive to both the richness of the action space and the availability of structured (hierarchical) training.

\subsubsection{HASAC: Stability and Consistency}
HASAC shows:
\begin{itemize}
  \item \emph{Stable positive performance} in cooperative, low-dimensional tasks (\textit{No Weapon}), reflecting the benefits of off-policy sample reuse and entropy regularization that promote both sample efficiency and smooth learning \citep{haarnoja2019softactorcriticalgorithmsapplications}.
  \item \emph{Competitive yields} in missile tasks, albeit with lower peak rewards than HAPPO.
\end{itemize}
This consistency reflects the benefits of off‐policy sample reuse and entropy regularization, which mitigate training instability in simpler coordination tasks.

\begin{table*}[!ht]
    \centering
    \begin{tabular}{|l|l|ccc|cc|}
        \hline
        \multirow{2}{*}{\textbf{Algorithm}} & \multirow{2}{*}{\textbf{env\_timestep}} & \multicolumn{3}{c|}{\textbf{No Weapon}} & \multicolumn{2}{c|}{\textbf{ShootMissile}} \\
        \cline{3-7}
        & & HierarchySelfplay & SelfPlay & vsBaseline & HierarchySelfplay & HierarchyVsBaseline \\
        \hline
        \multirow{3}{*}{HAPPO}
            & 510000  & -54.13 & -13.81 & -109.71 & 385.27 & 13.52 \\
            & 1020000 & -37.56 & -41.23 & -95.81  & 910.50 & 25.07 \\
            & 1500000 & 31.61  & -66.92 & -96.68  & 782.46 & 1090.17 \\
        \hline
        \multirow{3}{*}{HASAC}
            & 130000  & 30.14  & -24.88 & 31.53   & -6.79  & 4.77 \\
            & 370000  & 30.25  & -31.12 & 30.68   & 735.59 & 203.80 \\
            & 610000  & 30.20  & 7.82   & 29.38   & 918.70  & 468.34 \\
        \hline
    \end{tabular}
    \caption{Comparison of evaluation rewards across algorithms and scenarios.}
    \label{tab:happo_hasac_results}
\end{table*}

\subsection{Visualization Analysis under NoWeapon Setting}

\begin{figure}
    \centering
    \includegraphics[width=\linewidth]{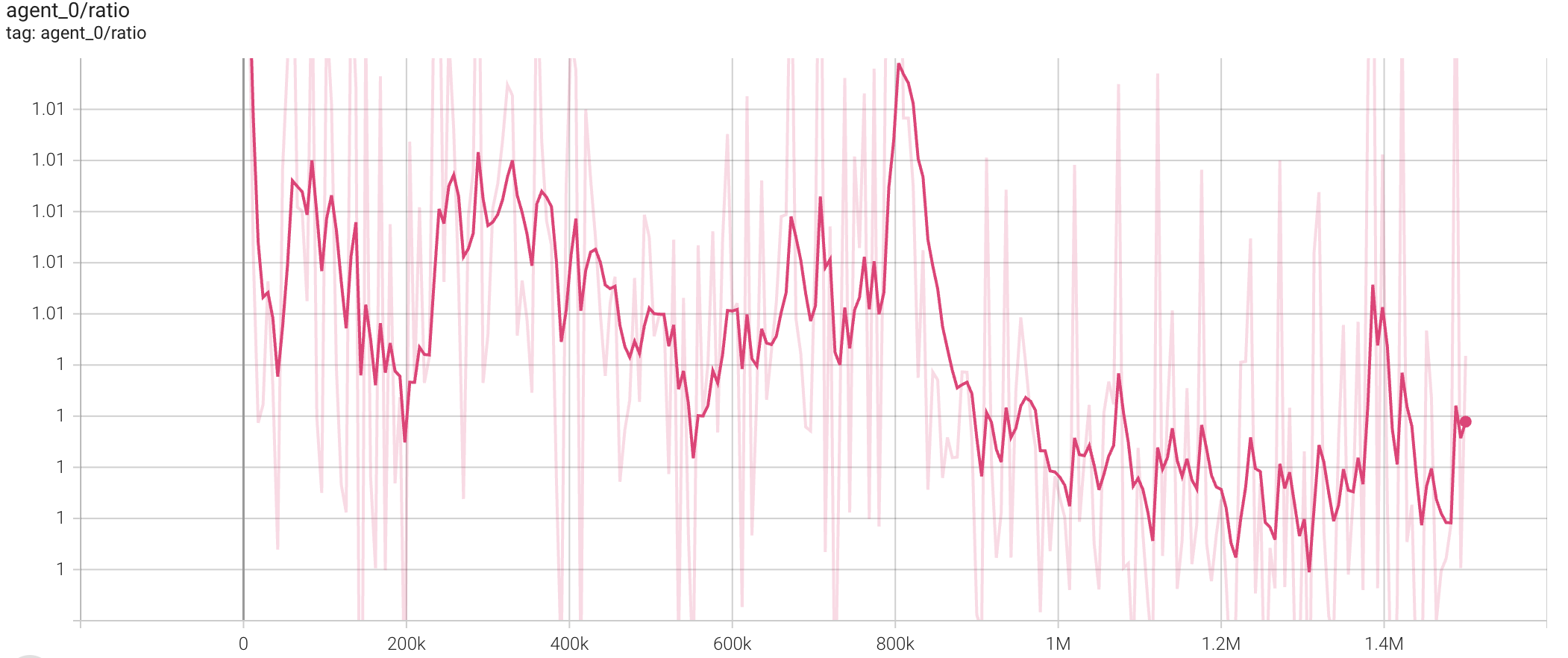}
    \caption{imp weights of agent of happo algorithms under NoWeapon, HierarchySelfplay experiment setting}
    \label{fig:1}
\end{figure}

\begin{figure}
    \centering
    \includegraphics[width=\linewidth]{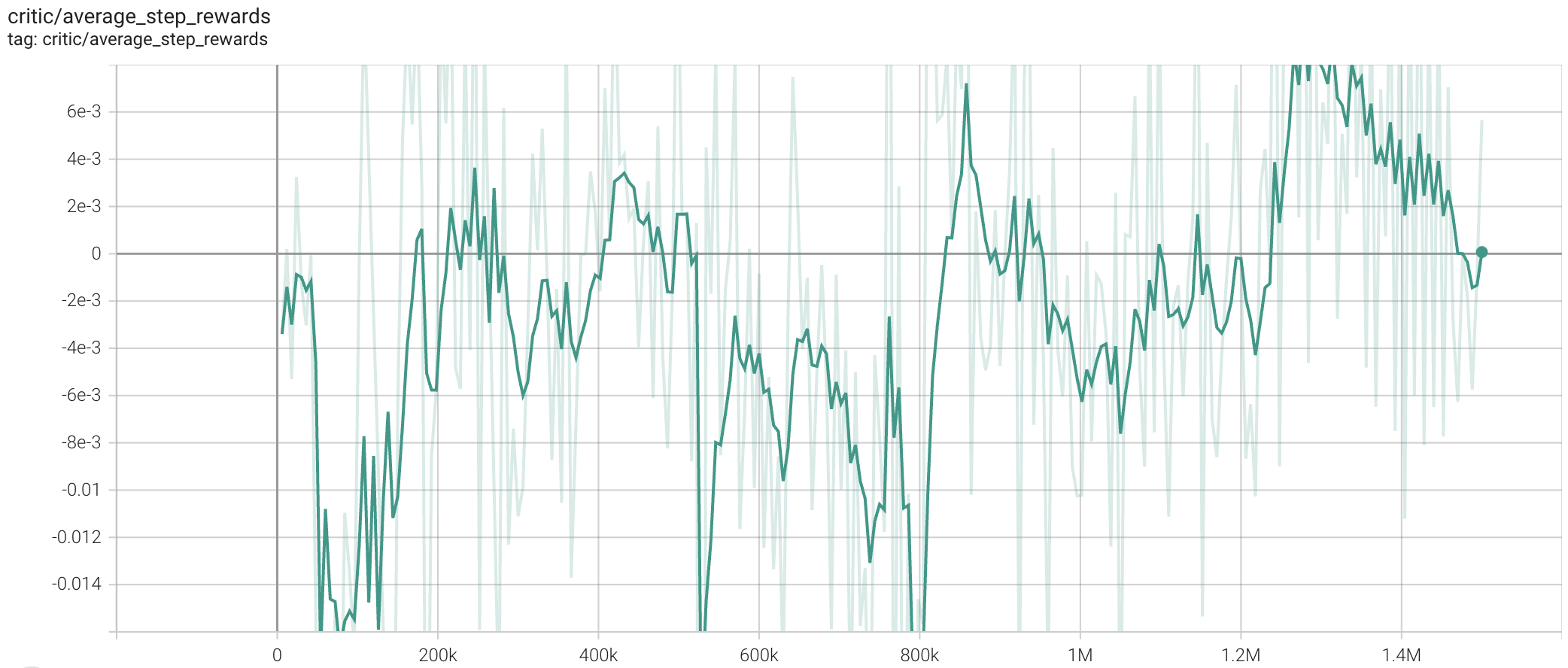}
    \caption{average step rewards of critic of happo algorithms under NoWeapon, HierarchySelfplay experiment setting}
    \label{fig:2}
\end{figure}

\begin{figure}
    \centering
    \includegraphics[width=\linewidth]{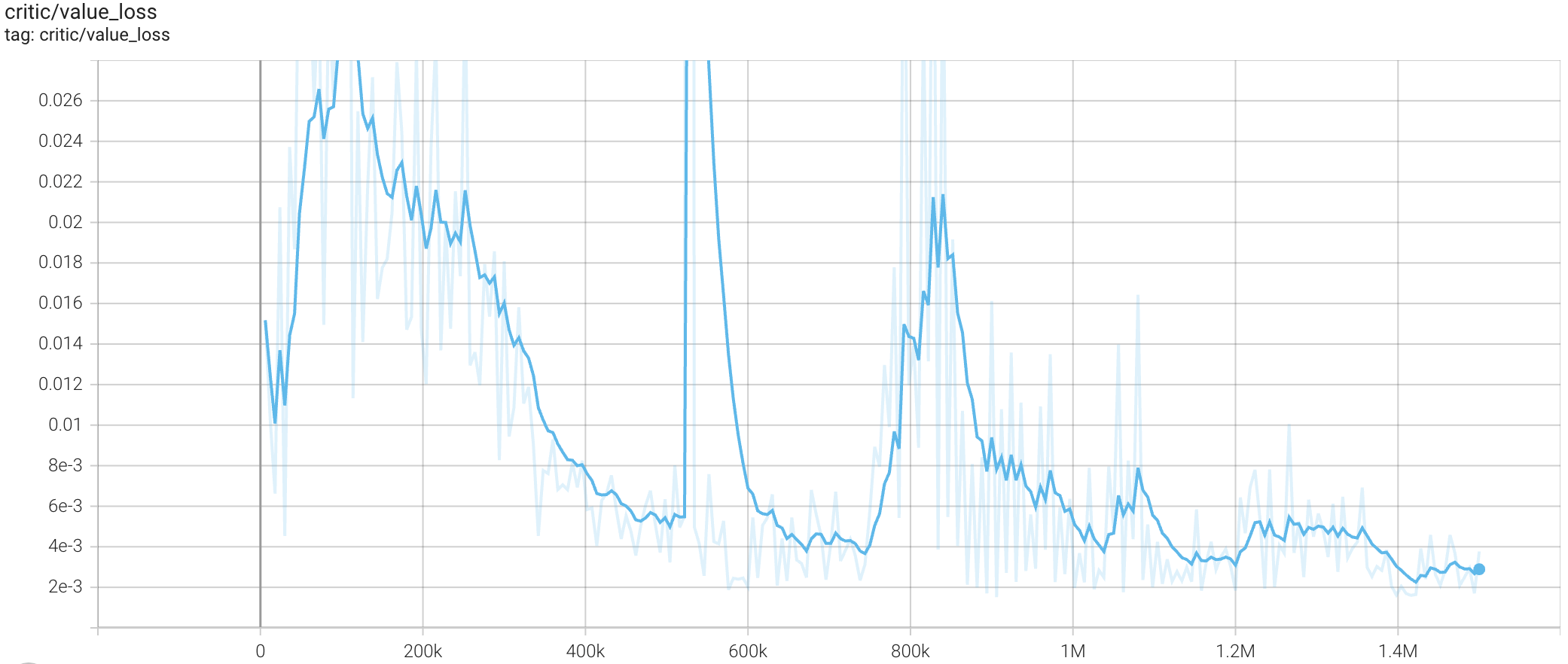}
    \caption{value loss of critic of happo algorithms under NoWeapon, HierarchySelfplay experiment setting}
    \label{fig:3}
\end{figure}

\begin{figure}
    \centering
    \includegraphics[width=\linewidth]{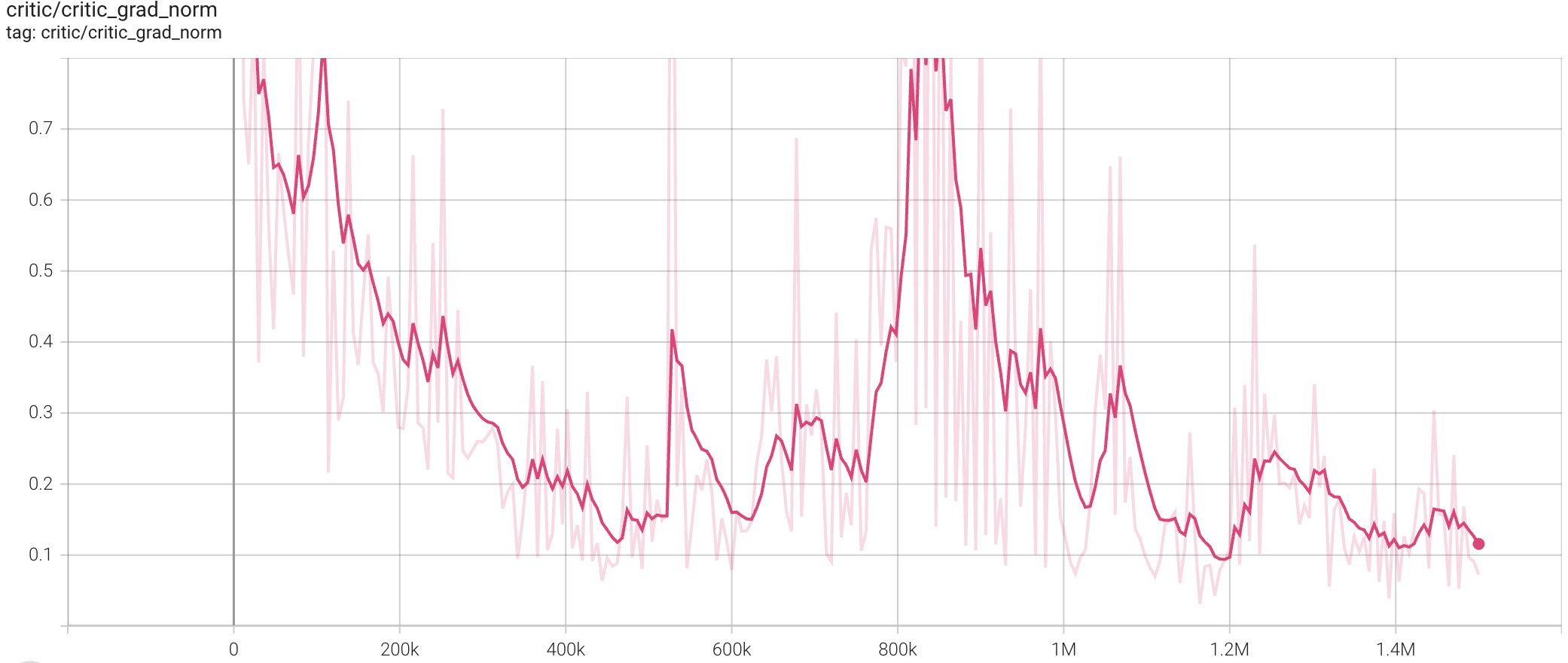}
    \caption{critic grad norm of agent of happo algorithms under NoWeapon, HierarchySelfplay experiment setting}
    \label{fig:4}
\end{figure}

\begin{figure}
    \centering
    \includegraphics[width=\linewidth]{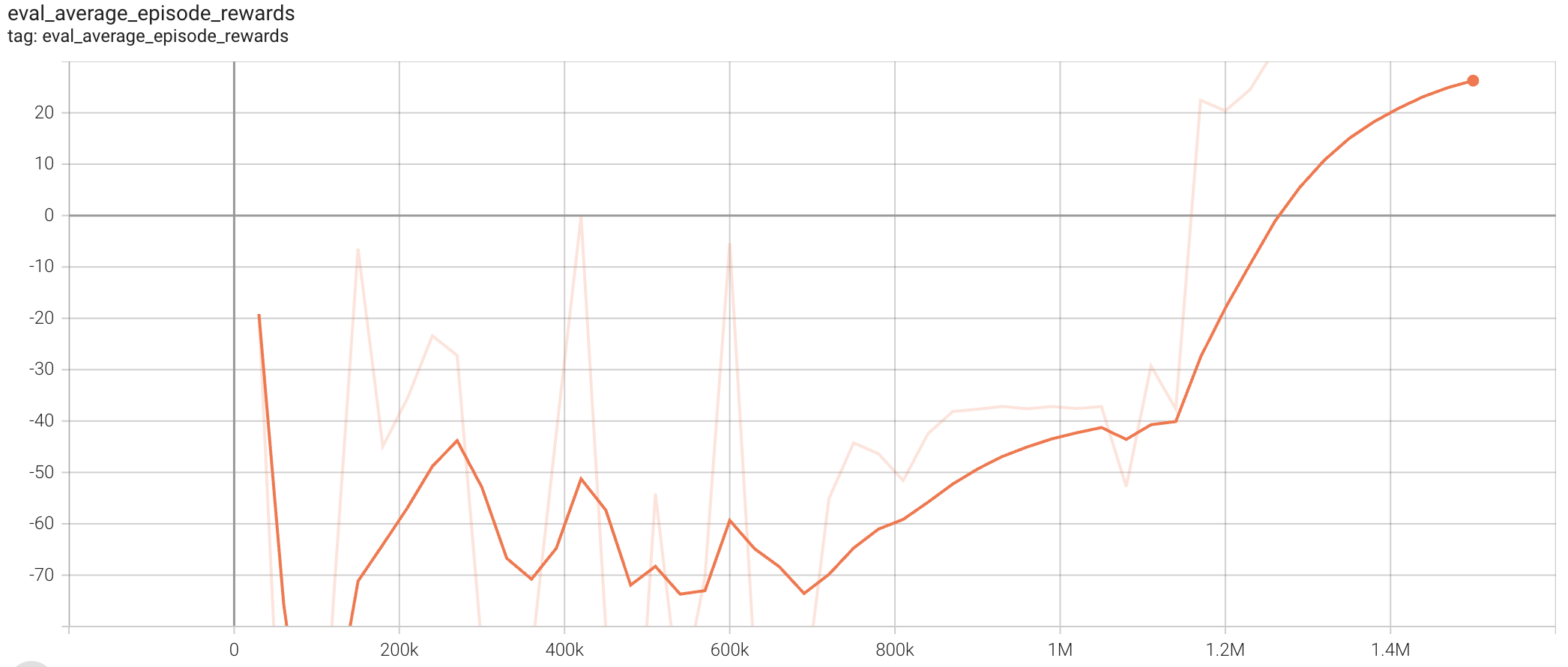}
    \caption{average episode rewrad in evaluation of happo algorithms under NoWeapon, HierarchySelfplay experiment setting}
    \label{fig:5}
\end{figure}

\begin{figure}
    \centering
    \includegraphics[width=\linewidth]{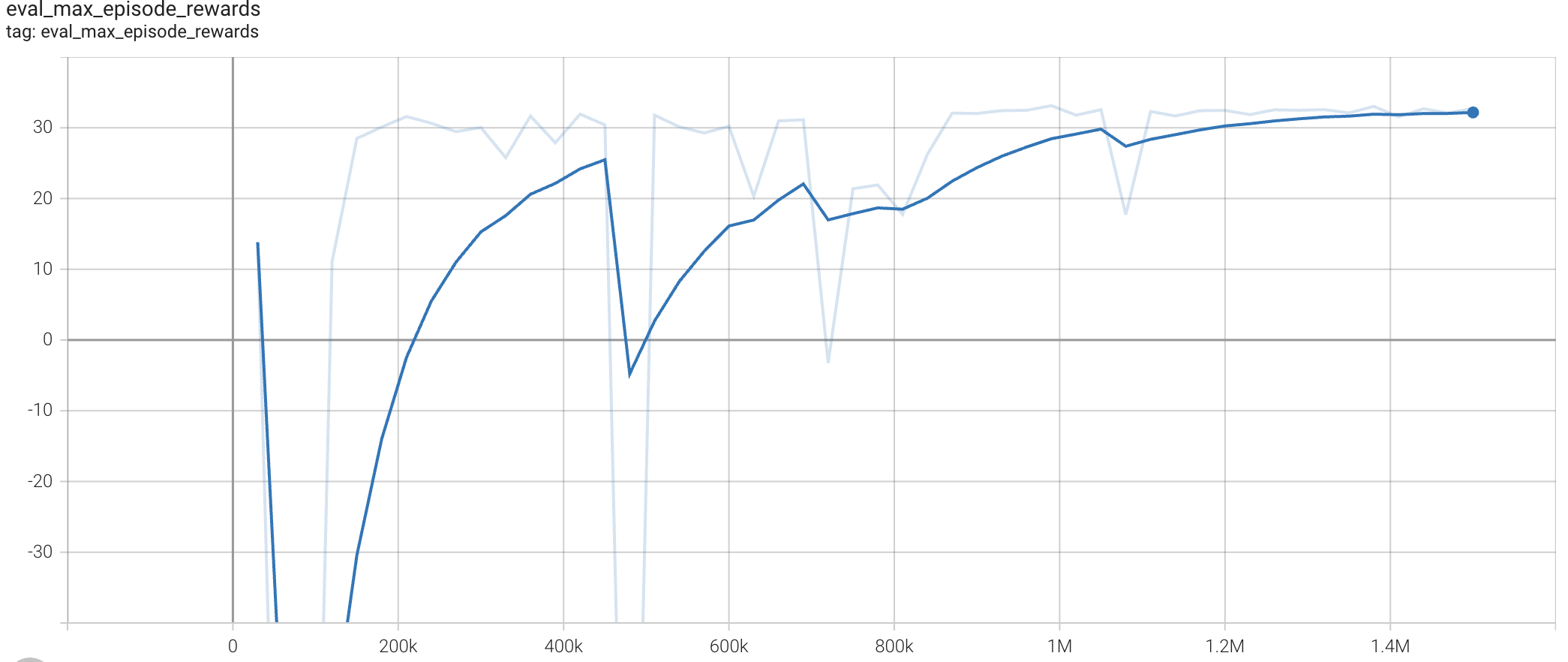}
    \caption{max episode reward in evaluation of happo algorithms under NoWeapon, HierarchySelfplay experiment setting}
    \label{fig:6}
\end{figure}

\begin{figure}
    \centering
    \includegraphics[width=\linewidth]{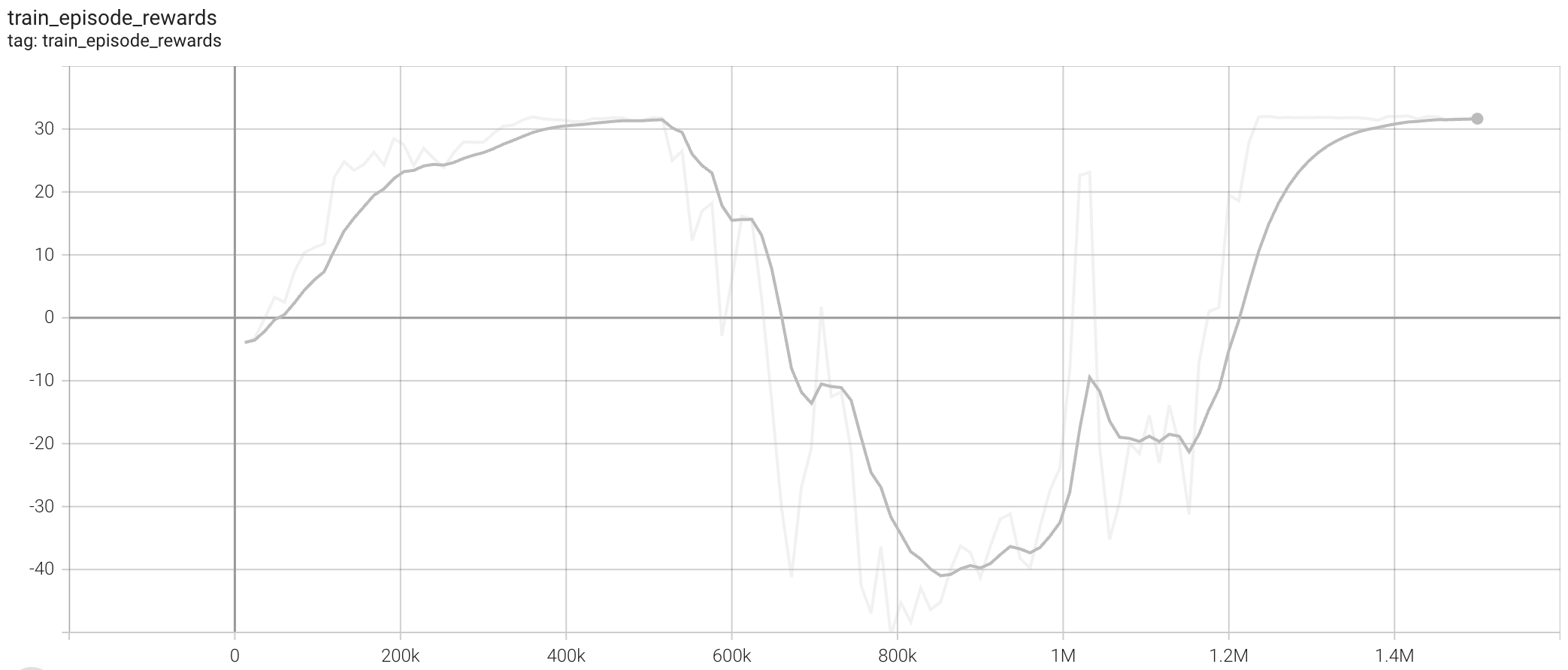}
    \caption{episode rewards in training of happo algorithms under NoWeapon, HierarchySelfplay experiment setting}
    \label{fig:7}
\end{figure}

\begin{figure}
    \centering
    \includegraphics[width=\linewidth]{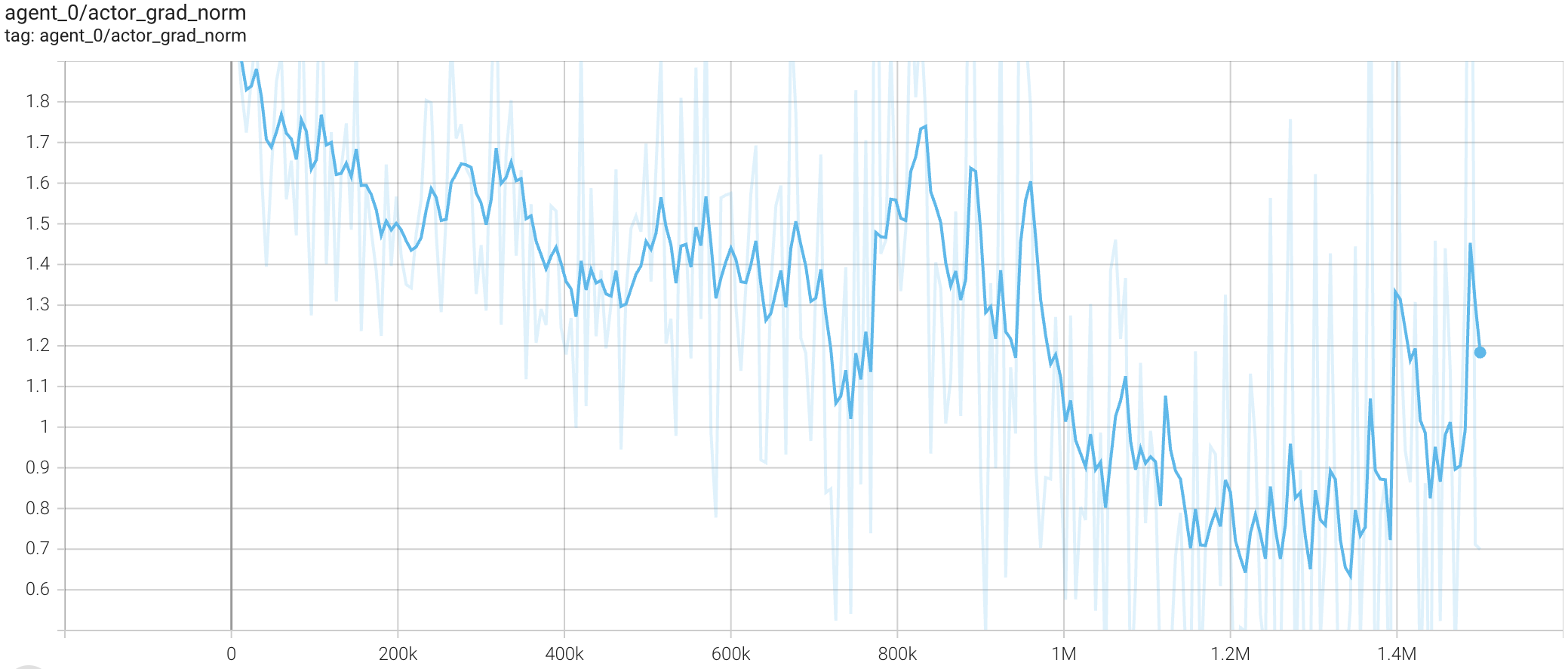}
    \caption{actors grad in training of happo algorithms under NoWeapon, HierarchySelfplay experiment setting}
    \label{fig:8}
\end{figure}

\begin{figure}
    \centering
    \includegraphics[width=\linewidth]{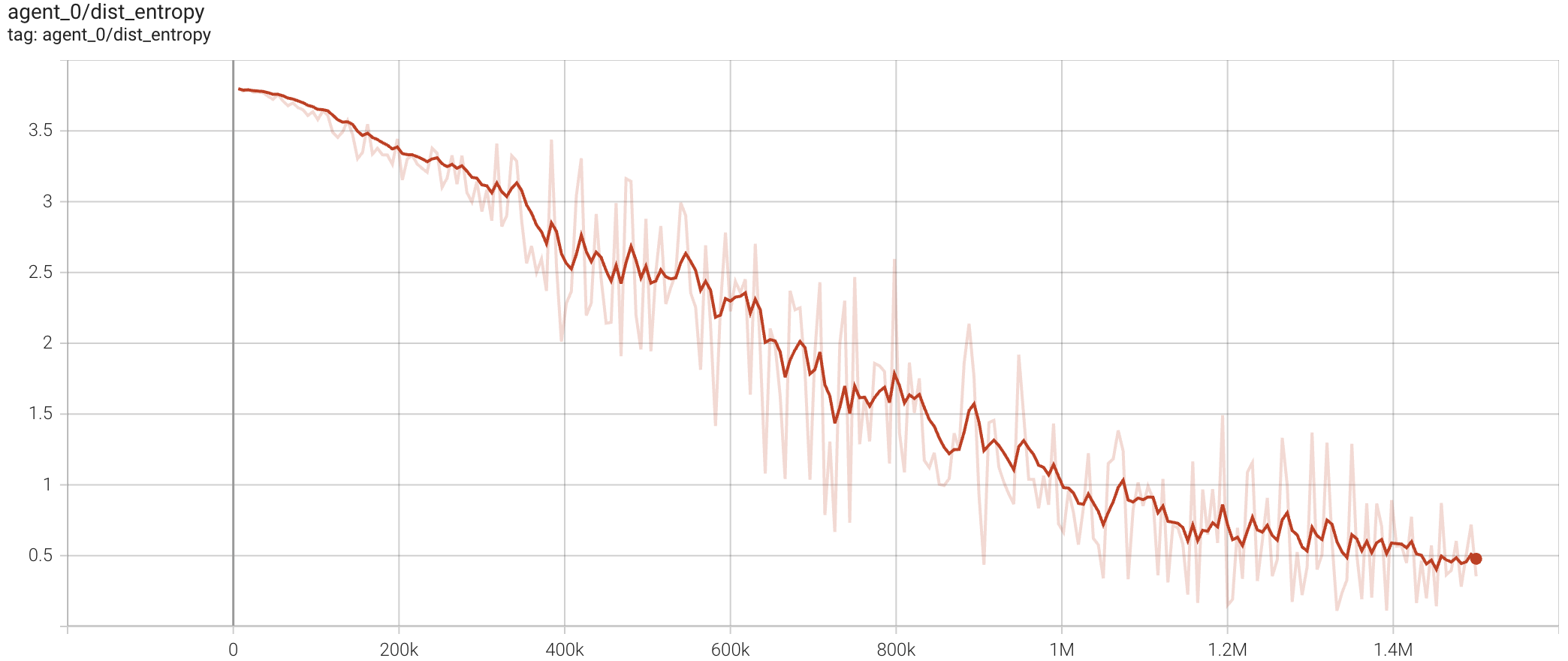}
    \caption{dist entropy of actors of happo algorithms under NoWeapon, HierarchySelfplay experiment setting}
    \label{fig:9}
\end{figure}

\begin{figure}
    \centering
    \includegraphics[width=\linewidth]{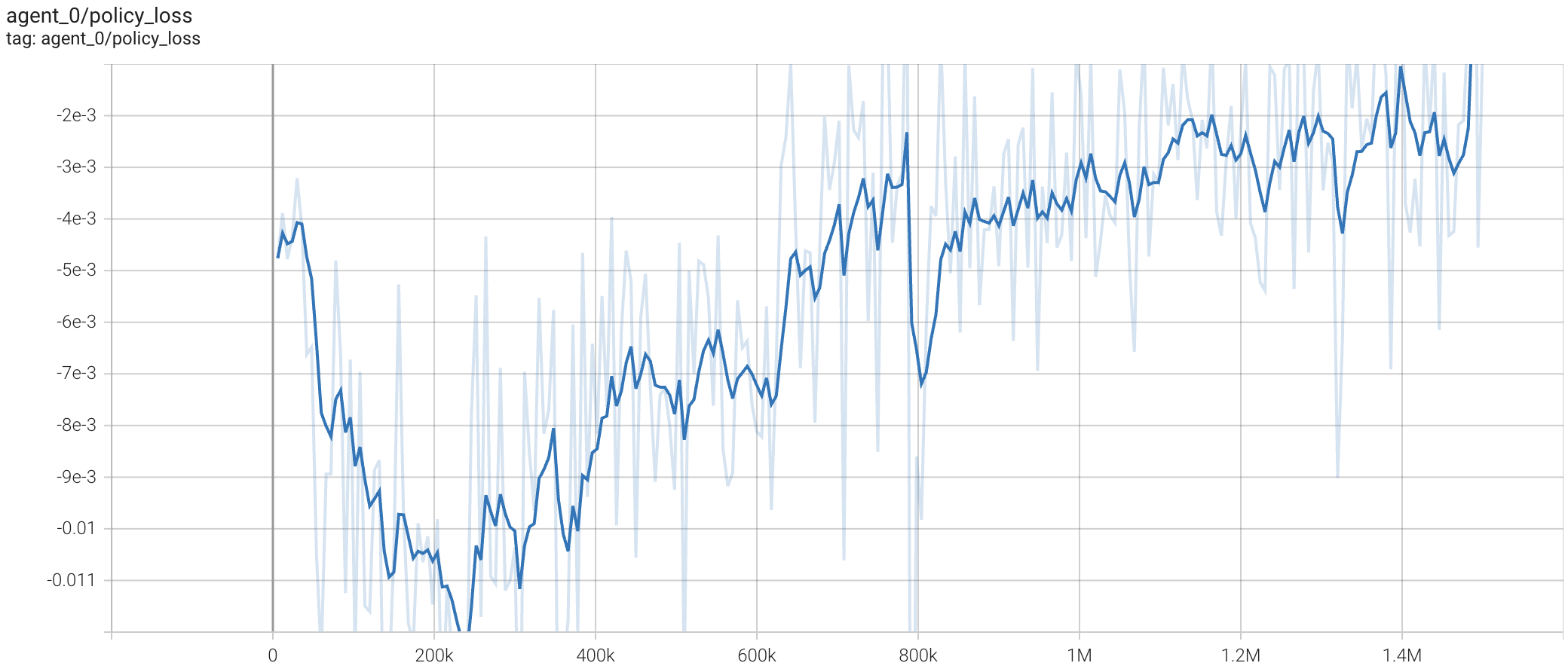}
    \caption{policy loss of actors of happo algorithms under NoWeapon, HierarchySelfplay experiment setting}
    \label{fig:10}
\end{figure}

In this section, we present visualizations of the training process using the HAPPO algorithm under the \texttt{NoWeapon} Setting 1. The results include ten plots, labeled from Fig.~~\ref{fig:1} to Fig.~~\ref{fig:10}, which illustrate various indicators collected throughout training: actor and critic metrics, evaluation-phase rewards, gradient magnitudes, and policy entropy.

\paragraph{Monitored Indicators.} The following quantities are visualized to assess the agent's learning behavior and training stability:

\begin{enumerate}
\item \textbf{Policy Loss:}
The clipped PPO surrogate loss that guides policy improvement, reflecting how well the updated policy aligns with the estimated advantage function.
\item \textbf{Dist. Entropy:}
The entropy of the action distribution, which promotes exploration and avoids premature convergence to deterministic suboptimal policies.
\item \textbf{Actor Grad Norm:}
The $\ell_2$-norm of the actor network's gradients (with optional clipping), serving as a measure of update magnitude and an indicator of training stability.
\item \textbf{Importance Weights:}
The policy likelihood ratio $r_t = \exp(\log\pi_{\theta}(a_t|s_t) - \log\pi_{\theta_{\text{old}}}(a_t|s_t))$, which quantifies the divergence between current and old policies and modulates the policy update strength.
\end{enumerate}

\paragraph{Training Reward Instability.}
From the reward curves in both training and evaluation (see Fig.~~\ref{fig:3} and Fig.~~\ref{fig:4}), we observe that the learning dynamics under HAPPO exhibit significant instability. The reward trajectory drops sharply multiple times during mid-training, indicating sensitivity to policy updates or value estimation errors. Such fluctuations suggest poor robustness of the algorithm in this environment, and the reward curve lacks smoothness, oscillating instead of showing stable monotonic improvement.

\paragraph{Symmetry Across Agents.}
Although we display results primarily for agent 0 due to space limitations, similar patterns are observed across all agents. The environment presents a high degree of symmetry, leading to largely mirrored training behaviors and metric trends. Therefore, visualizing one agent suffices to generalize insights to the full multi-agent system.

\paragraph{Critic Network Instability.}
A striking feature in the visualizations is the extreme ruggedness of the critic's training loss curve (see Fig.~~\ref{fig:2} and Fig.~~\ref{fig:3}), which contrasts sharply with the comparatively smoother actor-related metrics. Notably, the irregular critic behavior appears to coincide temporally with major drops in the evaluation reward, hinting at a causal relationship. This suggests that instability in value function estimation may propagate to the policy updates, resulting in erratic agent performance. Stabilizing the critic, possibly through better value targets or auxiliary objectives, could mitigate this issue.


\section{Conclusion}
\label{conclusion}

Through a series of controlled experiments in the LAG environment, we find that HASAC achieves stronger performance in static coordination tasks (\textit{No Weapon}), while HAPPO performs better in dynamic settings (\textit{ShootMissile}) that require richer exploration. Training diagnostics further reveal HAPPO's instability, likely due to critic fluctuations. This highlights a key trade-off: off-policy methods offer stable learning for basic interactions, whereas on-policy methods are better suited for complex, expressive behaviors—provided critic learning is well stabilized.

\end{document}